\documentclass[twocolumn]{rps-esrel2022}

\def\papername{\jobname}

\usepackage{natbib}
\usepackage{makecell} 

\begin{document}

\markboth{Jonk et al.}{Natural Language Processing of Aviation Occurrence Reports for Safety Management}

\twocolumn[

\title{Natural Language Processing of Aviation Occurrence Reports for Safety Management}

\author{Patrick Jonk, Vincent de Vries, Rombout Wever}

\address{Aerospace Operations Safety and Human Performance, Royal Netherlands Aerospace Centre, The Netherlands. \email{patrick.jonk@nlr.nl, vincent.de.vries@nlr.nl, rombout.wever@nlr.nl}}

\author{Georgios Sidiropoulos, Evangelos Kanoulas}

\address{Amsterdam Business School, University of Amsterdam, The Netherlands. \email{g.sidiropoulos@uva.nl, e.kanoulas@uva.nl}}

\begin{abstract} 
Occurrence reporting is a commonly used method in safety management systems to obtain insight in the prevalence of hazards and accident scenarios. In support of safety data analysis, reports are often categorized according to a taxonomy. However, the processing of the reports can require significant effort from safety analysts and a common problem is interrater variability in labeling processes. Also, in some cases, reports are not processed according to a taxonomy, or the taxonomy does not fully cover the contents of the documents. This paper explores various Natural Language Processing (NLP) methods to support the analysis of aviation safety occurrence reports. In particular, the problems studied are the automatic labeling of reports using a classification model, extracting the latent topics in a collection of texts using a topic model and the automatic generation of probable cause texts. Experimental results showed that (i) under the right conditions the labeling of occurrence reports can be effectively automated with a transformer-based classifier, (ii) topic modeling can be useful for finding the topics present in a collection of reports, and (iii) using a summarization model can be a promising direction for generating probable cause texts.

\end{abstract}

\keywords{Machine Learning, Natural Language Processing, Artificial Intelligence, Occurrence Reporting, Aviation, Safety Management.}

]

\section{Introduction}\label{introduction}

Occurrence reporting is a common method in safety management systems to obtain insight in the prevalence of hazards and accident scenarios, which enables safety analysts to identify hazards and to assess and mitigate risks. The International Civil Aviation Organization (ICAO) prescribes that service providers such as airlines, air traffic control organizations, ground handlers and airports report several categories of incidents and accidents when they occur. 
Reports are collected and investigated by independent organizations such as the National Transportation Safety Board (NTSB) of the United States and the Transportation Safety Board (TSB) of Canada.
In addition, many organizations collect and process reports outside the mandatory system. Examples of voluntarily reported occurrences are some of the Airline Safety Reports (ASR) that are collected by airlines as part of their safety management system and the NASA Aviation Safety Reporting System (ASRS) system.

As such, a great number of reports are collected each year. For example, the total intake of ASRS for the year 2020 was 65,656 reports \citep{asrs_program_briefing} and the Dutch Inspectie Leefomgeving en Transport (ILT) collected 13,802 reports in the year 2019 \citep{ilt}. In order to enable effective monitoring of risks and the extraction of useful insights from these reports, considerable effort is required to categorize and study the occurrences. Techniques from Natural Language Processing (NLP) may reduce the workload of processing the reports and may offer additional insights to safety analysts. This paper gives an overview of three research projects on NLP methods to support the analysis of safety occurrence reports. 

The first project discussed is the automatic labeling of reports using a classification model. In support of safety data analysis, reports are often categorized according to a taxonomy. However, the processing of the reports can require significant effort from safety analysts. In addition, inter-rater variability is a common problem in the labeling process. The automatic labeling of occurrence reports using a machine learning model may greatly reduce the effort of processing occurrence reports, improve the uniformity of the labels applied, and may offer additional insights in repositories where a labeling process is not yet part of the occurrence reporting system.

Traditional machine learning approaches such as support vector machines \citep{Tanguy} and random forests \citep{9049187} were used in the past for the automatic labeling of occurrence reports. The current study uses a Bidirectional Encoder Representations from Transformers (BERT)-based  classifier \citep{DBLP:journals/corr/abs-1810-04805}. BERT is a contextualized language model that holds state-of-the-art performance in various NLP tasks (e.g. question answering, sentiment analysis, etc.). Training and evaluation of the models was done using data from NASA's ASRS system and Canada's Civil Aviation Daily Occurrence Reporting System (CADORS).

NLP methods may enable the extraction of information and insights from occurrence reporting databases beyond the taxonomies that were applied by analysts to the reports in the repository. In particular, the goal of the second study is to find relations between different cause and effects present in the reports of the aviation accident database of the NTSB by their co-occurrence. In order to do this the latent topics in a collection of texts are extracted using a topic model. 

Latent Dirichlet Allocation (LDA) is a method from unsupervised machine learning that can be used for topic modeling, in which the topics from a collection of texts are extracted without the need for annotated data \citep{10.5555/944919.944937}.
Previous studies such as \citep{Tanguy} used LDA on occurrence reporting data from ASRS, where the applicability of the used methods was inconclusive. LDA allows the assignment of single documents to multiple topics, in contrast to unsupervised methods such as clustering. Assigning single documents to multiple topics is necessary for studying the co-occurrence of different factors.

The final project focuses on automatic summary generation using a language model. Like the labels discussed in the first project, reports can have many sections or attributes. Some of these attributes are free text sections. Again, because of the large number of reports that are annually collected, assisting in or automating some of the processing by a text generation model can reduce the effort required by safety analysts. One such free text attribute that can be automatically generated is the `probable cause and findings' section that is present in the NTSB reports, which is the topic of this study.

Automatic summarization of aviation occurrence reports is a topic that has not been studied extensively in the past. A similar sequence-to-sequence task in the aviation occurrence reporting domain that was studied before was question answering on ASRS reports \citep{asrs_question_answering}. 
This study used a variant of the previously discussed BERT model. However, the model architecture that was used in the current study was that of a Bidirectional and Auto-Regressive Transformers (BART)-large model  \citep{DBLP:journals/corr/abs-1910-13461}. 

Section \ref{methods} describes the models used in the study. In section \ref{data} the data are discussed, as well as the cleaning and processing steps required. The metrics used to quantify the model performance are discussed in \ref{metrics}, after which the metrics calculated on the model predictions and some examples of predictions are shown in section \ref{results}. Additional details regarding the data sets, hyperparameters, baseline models and example results are available at github.com/AviationNLP/nlp\_aviation\_safety.

\section{Methods}\label{methods}
\subsection{BERT}
A BERT-based approach was used for the classification models\footnote{Several neural and non-neural approaches were tested, but the BERT-based model held the best performance in preliminary experiments.} \citep{DBLP:journals/corr/abs-1810-04805}.
BERT makes use of the encoder part of the transformers architecture. One of the reasons that transformer models are successful is that they produce contextual representations where the meaning to the model of the input words is dependent on the context in which they appear (word-sense disambiguation).  Transformers make use of self-attention, which is a mechanism that allows the models to understand relations between words that are spaced an arbitrary distance apart in a text.
In addition, the model uses transfer learning. Transfer learning is a paradigm where a model is first trained on a task for which a large amount of data is available. Subsequently, the model is fine-tuned on the task of interest. The pre-training task is similar enough to the task of interest for the model to learn valuable information during the pre-training phase. 

In text classification problems a model is trained to assign one or more categories where the input is a sequence of words. One of the preprocessing steps when using a BERT-based model is prepending the input sequence with a CLS token. For classification tasks a fully connected layer is added to the model output of the CLS token. The dimension of the output vector of the last layer is equal to the number of possible output categories (e.g., if we have 7 human factors as categories, the last layer would have been of size 7). This study focuses on multi-label classification problems, meaning that one or more labels are assigned to each document. The sigmoid activation function is applied to the outputs, casting them to independent probability scores between zero and one, as opposed to the softmax activation function in single-label classification where the sum of the output scores is normalized. Typically a threshold is applied to the probability scores to decide whether the corresponding label is applicable.

\subsection{LDA}
LDA topic models are mainly used to discover hidden topics in a corpus, but they can also be used to classify documents. The output of the model is a list of topics of length $n_{topics}$, where each topic in the corpus consists of a probability distribution over all possible words. In addition, for each document a list of length $n_{topics}$ is given, containing probability scores indicating to what degree each topic belongs to the given document. Intuitively, words that often co-occur in documents represent underlying topics in the corpus. 

LDA is a generative probabilistic topic model. A number of assumptions are made about how documents are generated. For each document:

\begin{romanlist}[(iii)]
\item{} The number of words $N$ is chosen from a Poisson distribution.
\item{} The topic distribution $\theta$ of a document is drawn from a Dirichlet distribution.
\item{} For each of the $N$ words in the document:
\begin{alphlist}[(b)]
\item  a topic $z$ is drawn from $\theta$
\item  each topic has a probability distribution over the words in the corpus. A word is drawn from the probability distribution belonging to $z$. 
\end{alphlist}
\end{romanlist}

During the training or inference phase the distribution of the topics over the corpus and the word distributions over the corpus are learned. LDA models do not offer control over the contents of the topics that will result from the model. Moreover, the number of topics is a hyperparameter of the model and there is no definitive method for determining the optimal number of topics. Based on a causal safety model of \citep{causal_model}, a total number of 40 topics was estimated to be a reasonable number.

\subsection{BART}
Finally, the BART-large model  \citep{DBLP:journals/corr/abs-1910-13461} was used for the summarization task. Like the BERT model, the BART model makes use of a transformers architecture and transfer learning. However, this model makes use of an encoder-decoder architecture. 
 When used for question answering, BERT is an `extractive' model, meaning the output text of the model can only consist of a subset of the words present in the input text (span prediction). In constrast, BART is an `abstractive' model, which can generate arbitrary sequences regardless of the model input.  

\section{Data}\label{data}
\subsection{ASRS}
The occurrence classifier algorithm was trained and tested against two datasets en taxonomies. In both models the narrative of the occurrence reports served as the input of the model. The first model was trained to produce one or more human factors labels using data from NASA's ASRS system. ASRS is a database developed and maintained by NASA to collect and store voluntary occurrence reports from pilots, air traffic controllers, maintenance personnel and cabin crew. The filed reports are each processed by two of NASA's analysts, after which the documents are anonymized. In total there were 12 distinct human factors categories. Examples of human factors labels are `Confusion' and `Fatigue'. A copy of the ASRS database can be requested at NASA \footnote{https://asrs.arc.nasa.gov/search/requesting.html}.

For the experiment the data was divided into a training, validation and test set of 85\%, 5\% and 10\%, or 50,755, 2,672, and 5,937 samples respectively. The training set was used to train the model and learn the model parameters. The validation set was used to select the optimal model, and the test set was used to give an estimate of the performance of the model on data not previously seen by the model.

\subsection{CADORS}
The second model was trained to output one or more ICAO Accident/Incident Data Reporting (ADREP) occurrence categories using data from CADORS. CADORS is the Canadian national safety data reporting system. The purpose of the system is to collect information about operational occurrences within the Canadian National Civil Air Transportation System. It is used to for early identification of potential aviation hazards and system deficiencies. The reports in the CADORS system are labeled according to the ICAO ADREP taxonomy. One of the attributes of the reports that is labeled is the occurrence category. To each document one or more of 27 distinct occurrence categories are assigned. The CADORS database can be openly queried on their website \footnote{https://wwwapps.tc.gc.ca/saf-sec-sur/2/cadors-screaq/}.

Again, the dataset was split into a training, validation and test set of respectively 72\%, 8\% and 20\% or 152,562, 16,952, and 42,446 samples. Examples of some of the possible categories are `Bird strike' and `Runway incursion'.

\subsection{NTSB Aviation Accident Database}
The NTSB is the agency of the United States government that investigates and reports on aviation accidents and incidents. Accident and incident reports are stored in the NTSB aviation accident database. The time frame of most of the accident and incident reports in the database available for this study spans from 1989 until 2019. The reports in the NTSB aviation accident database are typically one or two pages long and contain a number of free text sections and some additional properties, some of which categorical. In general, the free text fields consist of three sections. These sections are `analysis’, `factual information’ and `probable cause and findings’. The contents of the NTSB database are openly available \footnote{https://data.ntsb.gov/avdata}. 

The `factual information' section contains general information about the flight. The `analysis' section provides an analysis of the facts and events that are directly relevant for the occurrence.  In the ‘probable cause and findings’ section the NTSB summarises the probable cause and contributing factors of the occurrence in the form of a one- or two sentence description. An example of a probable cause and findings section is: ``The pilot's loss of control while on approach to the runway. Contributing factors were the downdraft and the lack of suitable terrain for the off-airport landing.''

After a number of preprocessing steps, the size of the data set to which the topic model was applied was 54,410 documents. During preprocessing duplicate reports were removed. Reports where the ‘probable cause and findings’ section was missing, as well as reports where both the ‘analysis’ and ‘factual information’ were missing were discarded. In addition, all characters were cast to lower case. The probable cause and findings sections were relatively short, since the average number of words is 25.3. In addition, the used vocabulary was found to be relatively homogeneous and domain specific.

In the case of the probable cause summarizer, texts that consisted of only uppercase characters were cast to lowercase since the model is case-sensitive. The fact that the report structure consists of an ‘analysis’, ‘factual information’ and ‘probable cause and findings’ section, and that the information in the ‘probable cause and findings’ is based on information in the first two sections, was used to define a summarization task. The ‘analysis’ and ‘factual information’ sections were concatenated and used as the input text. The model was then trained to generate the ‘probable cause and findings’ section as the target text.

Because the time and memory available for computation was limited the maximal length of the input text was set to 1024 tokens. All reports with an input text larger than 1024 tokens, in total 11,945 reports, were discarded, after which 42,466 reports were left. The cleaned data set was split up into a training, validation and test set of 70\%, 15\% and 15\%, or 29,726, 6,370 and 6,370 reports respectively.

\section{Metrics}\label{metrics}
In order to quantify the performance of the models a number of metrics were calculated. For the occurrence classifier, these metrics were the precision $P$, recall $R$, success $S$ and exact match $EM$. Using the true positives $TP$, false positives $FP$, and false negatives $FN$, precision and recall are defined as $P = TP/(TP + FP)$ and $R = TP/(TP + FN)$. The success score is calculated by dividing the number of documents to which at least one of the ground-truth labels is assigned by the total number of documents.

The output of the model is a list of categories with associated probabilities. A higher probability score indicates a higher confidence the category is applicable to the document. When the labels per topic are ordered by probability score, $P@n$, $R@n$ and $S@n$ can be found by only taking into account the top n most probable assigned categories when calculating these metrics.

The $EM$ score is calculated by dividing the number of documents to which all the labels are correctly assigned by the total number of documents. In the classification task, the predicted labels are the labels with a probability higher than the threshold of 0.5. For the topic modeling study, only $P$ and $R$ have been calculated.

ROUGE (Recall-Oriented Understudy for Gisting Evaluation) is a set of metrics that is often used to quantify the quality of automatically generated summaries or translations \citep{lin-2004-rouge}. The $R1$, $R2$ and $RL$ precision and recall scores were calculated for the texts generated in the test set. $R1$ and $R2$ scores are the fractions of uni- and bigrams that overlap respectively between the generated text and the existing reference `probable cause and findings' section from the report. In case of the precision, the denominator is the number of N-grams in the generated text. In case of recall, the denominator is the number of N-grams in the existing reference text. For example the $R1$ recall would be calculated as $R1 = (\text{\textit{unigrams in both}})/(\text{\textit{unigrams in reference}})$.

N-grams are sequences of N consecutive words in a text. For example, for ‘excessive breaking caused the accident’, the unigrams are ‘excessive’, ‘breaking’, ‘caused’, ‘the’ and ‘accident’. The bigrams are ‘excessive breaking’, ‘breaking caused’, ‘caused the’ and ‘the accident’, etc.

The $RL$ score refers to the fraction of the length of the longest common sub-sequence between the generated and existing reference text, divided by the total number of unigrams in the generated or reference text.

\section{Results and Discussion}\label{results}
\subsection{Occurrence Classifier}\label{results_occurrence_classifier}
The performance metrics of both the human factors and the occurrence category classifier are shown in table \ref{tab4}. The $S@1$, $S@2$ and $S@3$ scores are 67\%, 80\%, and 92\% respectively, thereby illustrating that for most cases at least one of the ground truth categories was found within the first two retrieved ones. The $R@1$ and $R@2$ are relatively low. This was expected, because many records had more than two ground truth labels.

\begin{table}
\tbl{Performance metrics of the BERT-based report categorization model.\label{tab4}}
{\tabcolsep14pt
\begin{tabular}{@{}lll@{}}\toprule 
Metric & \makecell{ASRS \\ (human \\ factors)} & \makecell{CADORS \\ (occurrence \\ category)} \\ \colrule
P@1 & 0.67 & 0.88\\  
P@2 & 0.56 & 0.56\\
P@5 & 0.37 & 0.24\\ \hline 
R@1 & 0.34 & 0.77\\  
R@2 & 0.53 & 0.92\\
R@5 & 0.81 & 0.98\\ \hline
S@1 & 0.67 & 0.88\\  
S@2 & 0.80 & 0.95\\
S@5 & 0.92 & 0.99\\ \hline
EM & 0.20 & 0.70\\ \botrule
\end{tabular}}
\end{table}

The $EM$ score of 20\% was relatively low. Since the data set was imbalanced across the different categories a relatively low performance was expected. However, when only categories are considered for which more than 1000 samples exist, the $EM$ only improves to around 30\%. A likely cause for the low performance is a low inter-rater agreement in the annotated data. This can negatively influence the training process and adds a degree of randomness to the ground truth labels in the test set, thereby lowering performance metrics.

The performance of the occurrence classifier trained on the CADORS data was significantly higher. The $EM$ score that was achieved was 70\%. The difference in performance between the human factors classifier and the occurrence category classifier can be due to a number of reasons. First, it is interesting to note that there were more distinct occurrence category labels, 27 in total, as compared to 12 human factors labels. One could think that fewer distinct labels make the problem easier to solve. However, to each of the documents of the ASRS data set on average a larger number of labels was assigned, therefore making it harder to predict all correct labels.

Other reasons for the difference in performance could be quality of the annotations. The reports from ASRS are gathered through self-reporting, whereas 80\% of the reports from CADORS come from ANSP NAV CANADA. The remaining of the reports is provided by TSB Canada, airports, etc. In addition, a less clear distinction between the labels defined in the taxonomy may cause the human factors classification task to be more complex than the occurrence category classification.

\subsection{Topic Model}
Interpreting what concept is represented by a certain topic can be done by examining the top $n$ most probable words in the probability distribution of a given topic. In this study a topic was defined to be of `good' quality if a single coherent concept could be deduced by a subject matter expert or safety analyst from the top 10 topic words, and `bad' quality if it could not. Analysis of the top 10 topic words of all topics showed that 32 out of 40 topics were of `good' quality. A selection of the `good' quality topics and the top topic words are shown in table \ref{tab1}. Note however that the interpretation of topics is quite subjective.

\begin{table}
\tbl{Selection of 9 of the good quality topics with description and top 10 words found by the LDA model. \label{tab1}}
{\tabcolsep8pt
\begin{tabular}{c c}\toprule 
Description & Top 10 topic words\\  \colrule
\makecell{landing \\ incident} & \makecell{go, around, runway, tailwind, \\ touchdown, anomaly, landing, \\ proper, point, attempted} \\ \hline
\makecell{landing \\ gear \\ damage} & \makecell{landing, gear, hard, nose, \\ main, right, left, \\ collapse, damage, resulting} \\ \hline
\makecell{instructor- \\ student} & \makecell{flight, instructor, action, delayed,\\ remedial, student, inadequate,\\ non, input, hover} \\ \hline
\makecell{planning \\ decision \\ making} & \makecell{decision, improper, his, \\ attempt, planning, off, \\ flight, land, take, tailwind} \\ \hline
\makecell{collision} & \makecell{which, resulted, an, with,\\ subsequent, airplane, flight,\\ collision, loss, impact} \\ \hline
\makecell{flare \\ bounced \\ landing} & \makecell{landing, improper, flare, \\ student, bounced, recovery, from,\\inadequate, supervision, misjudged} \\ \hline
\makecell{loss \\ of \\ control} & \makecell{during, control, maintain,\\landing, directional, airplane,\\roll, loss, takeoff, aircraft} \\ \hline
\makecell{propellor / \\ crankshaft \\ failure} & \makecell{fatigue, due, propeller, blade,\\engine, separation, from,\\fracture, bearing, crankshaft} \\ \hline
\makecell{loss \\ engine \\ power} & \makecell{engine, for, reason, loss,\\power, undetermined, carburetor,\\total, partial, due} \\ \botrule
\end{tabular}}
\end{table}

After training the model, 65 documents were randomly sampled, after which the topics of each document were sorted by probability. The validity of the top 3 most probable topics when compared to the content of the document was evaluated by the NLR team. Subsequently, the precision and success were calculated for the top 3 topics. The precision and success scores are shown in table \ref{tab2}.

\begin{table}
\tbl{Precision and success scores of the LDA model that was fit on NTSB data, calculated from 65 randomly sampled reports.\label{tab2}}
{\tabcolsep14pt
\begin{tabular}{@{}lll@{}}\toprule 
Topics evaluated & Precision & Success\\ \colrule
top 1 & 70.8 & 70.8\\  
top 2 & 56.2 & 83.1\\
top 3 & 46.2 & 87.7\\ \botrule
\end{tabular}}
\end{table}

As seen in table \ref{tab2} $P@1$, $P@2$, and $P@3$ were found to be 70.8\%, 56.2\%, and 46.2\% respectively. Arguably, the precision @ 1 might be acceptable for some applications, but for the precision @ 2 and 3 the performance is poor for practical applications. Therefore this model might not be suitable for inferring relations between different causes and between causes and effects.

$S@1$, $S@2$ and $S@3$ were found to be 70.8\%, 83.1\% and 87.7\%. When the top 3 topics are taken into account, a considerable fraction of the documents has at least one correct topic assigned. For applications in which false positives are of low importance this model might be suitable.

It should be noted that the sample size of the evaluated documents was rather small and may not provide a statistically robust score. In addition, the precision and success scores were calculated across all topics. It is possible that significant variance exists in the performance of the model for the different topics. In short, the lack of test data makes it challenging to quantify the quality of the predictions at this time. 

\subsection{Probable Cause Summarizer}
Below are some examples of reference ‘probable cause and findings’ sections from the reports in the test set, and the text generated by the model. 

(1 - reference) The pilots failure to maintain directional control during landing roll that resulted in a collision with an airport sign. Factors were wet runway surface and wind gust.

(1 - generated) The pilot's failure to maintain directional control during landing. A factor was the wind gust.

(2 - reference) Failure of the flight control lower mixing unit due to fatigue, which resulted in the loss of control.

(2 - generated) A fatigue failure of the lower mixing unit. contributing to the accident was unsuitable terrain.

(3- reference) A loss of left braking action for reasons that could not be determined because postaccident examination did not reveal any evidence of preimpact mechanical failures or malfunctions that would have precluded normal operation, which resulted in a loss of directional control.

(3 - generated) The pilot’s inability to maintain directional control during the landing roll due to a lack of left brake action for reasons that could not be determined.

(4 - reference) failure of the number three exhaust valve on the left engine. a major factor was the high density altitude where the airplane would not operate on single engine.

(4 - generated) The failure of the number three exhaust valve. a factor in the accident was the high density altitude.

The examples above show that the model generates grammatically correct language that is often indiscernible from human written text. The generated texts were not always correct however with regards to the content. For example, in example 1 `wet runway surface' is stated as one of the factors in the reference text but not in the generated text. The ROUGE scores give an estimate of the agreement between the generated and reference texts. The $R1$, $R2$ and $RL$ precision and recall calculated on the test set are shown in table \ref{tab3}.

\begin{table}
\tbl{ROUGE scores of the BART-based model trained on NTSB reports, on texts from the test set.\label{tab3}}
{\tabcolsep14pt
\begin{tabular}{@{}lll@{}}\toprule 
Score & Precision & Recall\\ \colrule
R1 & 59.7 & 49.4\\  
R2 & 39.0 & 32.1\\
RL & 52.8 & 43.8\\ \botrule
\end{tabular}}
\end{table}

As compared to the performance of the BART model on the CNN/DailyMail dataset \citep{DBLP:conf/nips/HermannKGEKSB15}, which is a standard dataset on which summarization models are often tested, the performance of the model in the current study is relatively high \citep{DBLP:journals/corr/abs-1910-13461}. On the CNN/DailyMail dataset the BART model achieves $R1$, $R2$ en $RL$ recall scores of 44.2\%, 21.3\% and 40.9\% respectively, as compared to 49.4\%, 32.1\% and 43.8\% for the NTSB dataset. The reason the model achieves a higher score in this study is probably related to the fact that the NTSB's reports are relatively homogeneous in language, sentence construction, phraseology, and that many reports contain similar contents with regards to scenario's and causes.

\section{Conclusion}\label{conclusion_and_discussion}
The labeling of occurrence reports can be effectively automated under certain conditions. The occurrence category model predicted the labels of the documents in the test set with a high degree of correctness, whereas the performance of the human factors model was significantly lower. As the number of applicable labels for each document increases the classification task becomes harder.
High quality, uniform annotations in the training data are necessary for high quality predictions. In addition, clear distinctly defined categories may improve the performance. 

Topic modeling is a technique that is suitable for an exploratory analysis of a set of documents to infer the main topics underlying the data set. In the context of aviation safety the application may be useful for occurrence reports or for the open-ended questions in a survey. However, because of the relatively low performance in assigning the correct topics to the different documents this model is not suitable for inferring relations between different causes and between causes and effects in the NTSB reports.

Summarization models are promising as an aid in processing documents for a safety analyst. The generated texts were grammatically correct and most of the time not discernible from human generated texts. In addition, the model achieves relatively high ROUGE scores. However, these scores only gives a measure of the similarity between the generated and the reference text. Conclusively judging the performance of the model with regards to correctness and completeness of the probable cause summaries would require additional inspection of the results by domain experts.

When considering the three research projects in this paper on a whole, parallels can be drawn between the automatic labeling of reports and the automatic summary generation. Both methods could be used in conjunction to reduce the workload of analysts processing reports. After the analyst provides a narrative of the incident or accident, the models could be used to automatically determine the applicable labels and provide a summary stating the probable cause of the accident. Thereby it could also improve homogeneity over reports processed by different analysts.

Overall, NLP is promising for the processing of reports and the extraction of insights in aviation safety management. Factors that make the application of machine learning challenging in this domain are the the lack of quality and uniformity of the data, limited depth of information present in the documents, the complexity of the taxonomies applied, lack of uniformity and correctness in the processing and labeling of existing documents, or the presence of low quality annotated data that may corrupt possible training data.

\begin{acknowledgement}
Part of this research was done as part of the SAFEMODE project. This project has received funding from European Union's Horizon 2020 Research and Innovation Programme under Grant Agreement No 814961 but this document does not necessarily reflect the views of the European Commission.

NLR has conducted part of this research into the application of machine learning in support of safety analyses within the framework of the NLR research program for the Dutch government.

\renewcommand\theequation{A.\arabic{equation}}
\setcounter{equation}{0}

\bibliographystyle{chicago}
\bibliography{References}
\end{acknowledgement}

\end{document}